\documentclass{article}

\usepackage[final]{neurips_2020} 

\usepackage[utf8]{inputenc} 
\usepackage[T1]{fontenc}    
\usepackage{hyperref}       
\usepackage{url}            
\usepackage{booktabs}       
\usepackage{amsfonts}       
\usepackage{nicefrac}       
\usepackage{microtype}      
\usepackage{graphicx}
\title{Latent Compass: Creation by Navigation}

\author{%
  Sarah Schwettmann*\textsuperscript{1}\\
  \texttt{schwett@mit.edu} \\
  \And
  Hendrik Strobelt*\textsuperscript{2,3}\\
  \texttt{hendrik.strobelt@ibm.com} \\
  \And
  Mauro Martino\textsuperscript{2,3} \\
  \texttt{mmartino@us.ibm.com} \\

}

\begin{document}

\maketitle
\vspace*{-36pt}
\begin{center}\textsuperscript{1}MIT Brain and Cognitive Sciences, \textsuperscript{2}IBM Research,\textsuperscript{3}MIT-IBM Watson AI Lab\end{center}

\section{Motivation}

In Marius von Senden’s \textit{Space and Sight}, a newly sighted blind patient\footnote{\textit{Space and Sight} collects the first-hand visual experiences of congenitally blind patients who underwent experimental cataract surgery in the early 20\textsuperscript{th} century. } describes the experience of a corner as lemon-like, because corners “prick” sight like lemons prick the tongue [9]. \textit{Prickliness}, here, is a dimension in the feature space of sensory experience, an effect of the perceived on the perceiver that arises where the two interact. In the account of the newly sighted, an effect familiar from one interaction translates to a novel context. Perception serves as the vehicle for generalization, in that an effect shared across different experiences produces a concrete abstraction [3] grounded in those experiences. Cezanne and the post-impressionists, fluent in the language of experience translation, realized that the way to paint a concrete form that best reflected reality was to paint not what they saw, but what it was \textit{like} to see [7]. We envision a future of creation using AI where \textit{what it is like to see} is replicable, transferrable, manipulable – part of the artist’s palette that is both grounded in a particular context, and generalizable beyond it. 

An active line of research maps human-interpretable features onto directions in GAN latent space. Supervised and self-supervised approaches that search only for anticipated directions [4,6] or use off-the-shelf classifiers to drive image manipulation in embedding space [11] are limited in the variety of features they can uncover. Unsupervised approaches [8,10] that discover useful new directions show that the space of perceptually meaningful directions is nowhere close to being fully mapped. As this space is broad and full of creative potential, we want tools for direction discovery that capture the richness and generalizability of human perception. Our approach puts creators in the discovery loop during real-time tool use, in order to (i) identify directions that are perceptually meaningful to them, and (ii) generate interpretable image translations along those directions.

\section{Approach}

We define and implement a method that maps an \textit{experience} onto a \textit{latent direction} so one can repeat the experience by moving in that direction from any starting place. The mapping is wordless: a user explores concept space by browsing a pool of generated images, identifying images that capture a visual dimension of interest, and sorting them into two groups.  Based on this sorting, a linear SVM is trained to construct a separating hyperplane, and a normal to this hyperplane is identified as the direction $d$ capturing the corresponding feature of visual experience. Any image can then be transformed some amount $\lambda$ along this dimension by passing its transformed $z$ vector through the generator: $G(z + \lambda d)$. 

\begin{figure}
  \includegraphics[width=\linewidth]{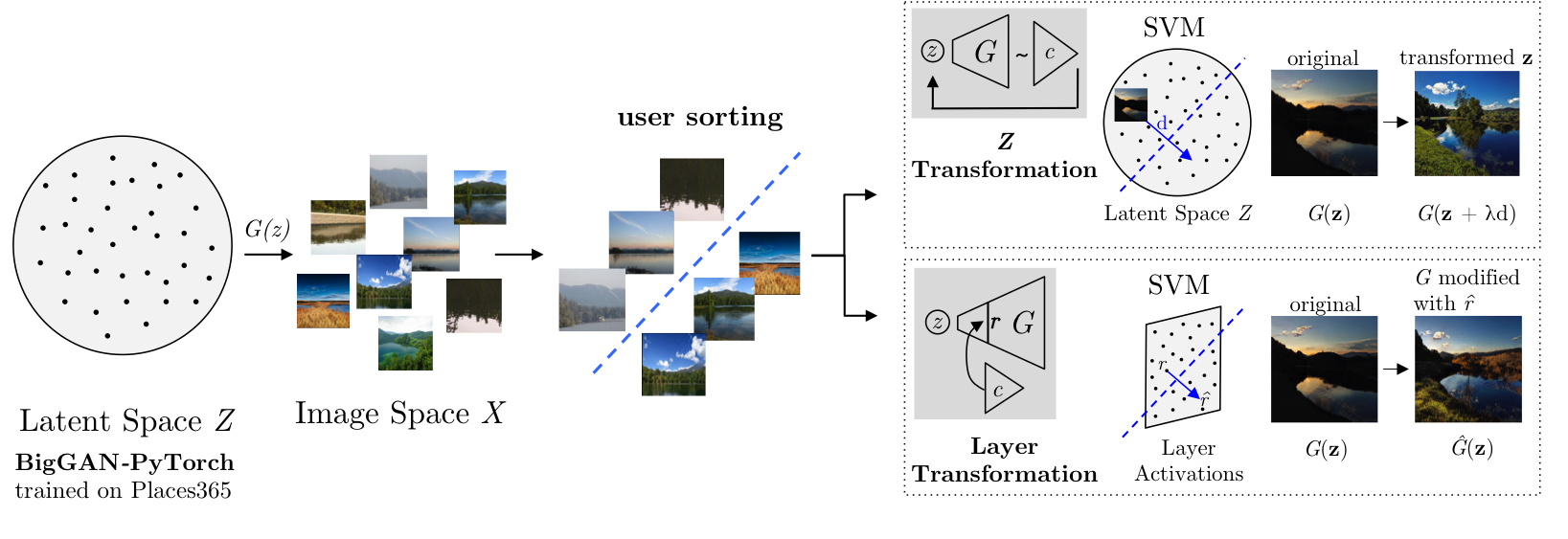}
  \caption{Schematic of direction discovery and image transformation. The method is broadly useful with different generators and could be augmented with an inverter to take real-world images as inputs.}
\end{figure}

In addition to using an SVM to drive changes in $z$, we can also intervene in intermediary layers and modify parts of the representation, differing from previous approaches which use the entire distribution modeling property of the generator. GAN dissection [1] has found that disentangled units in different layers of the generator control specific object classes across a variety of contexts, with increasing specificity for low-level detail in layers closer to the output. By training a classifier on unit activations in an individual layer, we can manipulate those activations along user-specified directions, producing image transformations that are more fine-grained than those in $Z$ space.

\section{Implementation}

The Latent Compass\footnote{Web demo: \href{latentcompass.com}{latentcompass.com} (best viewed on desktop/laptop). } is a tool for \textit{creation by navigation} in structured space. Users discover and travel along visual directions in chosen scenes. Our implementation uses BigGAN-PyTorch [2] trained on the Places365 dataset, which includes scenes from 365 unique categories [12]. Calibrating the compass reifies a perceptual experience, similarity, or difference into a transformation that can be applied to any image. To calibrate, users first select one of the Places categories to work in. Then they define two classes of images by moving examples of each to the left and right sides of the screen.\footnote{The tool requires a minimum of 14 images, roughly equal numbers in each class, for best results in training the underlying classifier.} Next, users choose whether to transform at the scene level ($Z$ manipulation) or at the detail level (Layer 1 manipulation). The compass is calibrated using the above method, using the classes to construct a direction that distills the difference between them: moving any image backward or forward along that direction transforms it into one class or the other. 

To navigate using the calibrated compass, users select a starting point by choosing a new image and moving it to the center of the map. Six transformations of the image are automatically generated, three steps forward and three steps backward along the trained direction. Arrows to the right and left let users move further in each direction. New trajectories can be added for comparison by dragging additional images onto the map. Directions can be labeled and saved after each training session.

\begin{figure}[h!]
  \includegraphics[width=\linewidth]{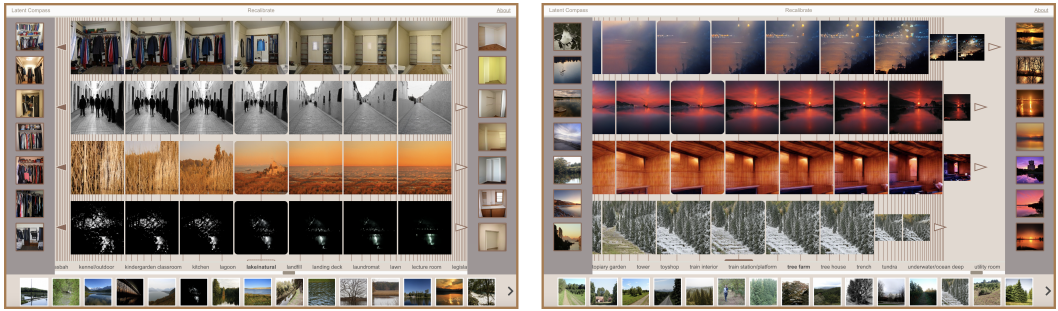}
  \caption{User interface showing two example transformations applied as scene-level changes (L) and fine-grained changes (R) across categories. The compass reveals that the "fullness" (L) and "gloaming" (R) directions are represented with sufficient abstraction to produce meaningful visual changes across categories. Attempting to pinpoint the perceptual decision boundary between two classes reveals the visual indeterminacy [5] characteristic of GAN imagery.}
\end{figure}

Navigation is not constrained to the calibration category. Users are invited to explore the same transformation in different scene categories to discover surprising translations of familiar experiences.

\section{Ethical Implications}
The dimensions that humans choose to explore and find interesting are reflective not only of their visual experience, but also of their biases.  We recognize the need to avoid drawing overly broad conclusions from the output of Compass users, as this is intended to be a tool for individual artistic exploration.  Furthermore, GANs themselves reflect the biases in their training data. While we do not have specific concerns about the Places365 dataset, our method is agnostic to the choice of GAN and would allow us to explore alternative models in the future. 

Direction labels will need light moderation before they are made public. A contact link on the site invites users to share feedback or report concerns.

\section*{References}

\small

[1] D. Bau, J.-Y. Zhu, H. Strobelt, B. Zhou, J. B. Tenenbaum, W. T. Freeman, and A. Torralba. GAN Dissection: Visualizing and Understanding Generative Adversarial Networks. In Proc. ICLR, 2019.

[2] A. Brock, J. Donahue, and K. Simonyan. Large Scale GAN Training for High Fidelity Natural Image Synthesis. In Proc. ICLR, 2019.

[3] P. Galison. (2012) Concrete Abstraction. In L. Dickerman (ed.), {\it Inventing Abstraction, 1910-1925: How a Radical Idea Changed Modern Art}, pp. 350-357. New York: MoMA. 

[4] L. Goetschalckx, A. Andonian, A. Oliva, and P. Isola. GANalyze: Toward Visual Definitions of Cognitive Image Properties. arXiv preprint arXiv:1906.10112, 2019.

[5] A. Hertzmann. Visual Indeterminacy in Generative Neural Art. In NeurIPS Creativity Workshop, 2019. 

[6] A. Jahanian, L. Chai, and P. Isola. On the "Steerability" of Generative Adversarial Networks. In Proc. ICLR, 2020.

[7] J. Lehrer. (2008) {\it Proust was a Neuroscientist}. Boston: Mariner Books. 

[8]  W. Peebles, J. Peebles, J.-Y. Zhu, A. Efros, and A. Torrralba. The Hessian Penalty: A Weak Prior for Unsupervised Disentanglement. In ECCV, 2020 (Spotlight).

[9] M. von Senden. (1960) {\it Space and Sight: The perception of space and shape in congenitally blind patients, before and after operation.} New York: Free Press.

[10] A. Vonyov and A. Babenko. Unsupervised Discovery of Interpretable Directions in the GAN Latent Space. arXiv preprint arXiv:2002.03754, 2020.

[11] C. Yang*, Y. Shen*, and B. Zhou. Semantic Hierarchy Emerges in Deep Generative Representations for Scene Synthesis. arXiv preprint arXiv:1911.09267, 2019.

[12] B. Zhou, A. Lapedriza, A. Khosla, A. Oliva, and A. Torralba. Places: A 10 Million Image Database for Scene Recognition. IEEE Transactions on Pattern Analysis and Machine Intelligence, 2017.

The authors would like to thank Samuel J. Klein for help crystallizing and communicating, and David Bau for elegantly mapping the landscape.

\end{document}